\documentclass{article}
\usepackage{amsmath}
\usepackage[T1]{fontenc}

\usepackage{csquotes}
\usepackage{xcolor}
\usepackage[inline]{enumitem}
\usepackage{graphicx} 

\usepackage[natbib=true,style=authoryear ]{biblatex}
\usepackage{tabularx} 

\title{Factors That Support Grounded Responses in LLM Conversations: A Rapid Review}
\author{Gabriele Cesar Iwashima, Claudia Susie Rodrigues, \and
Claudio Dipolitto and Geraldo Xexéo}
\date{June 2025}

\begin{document}

\maketitle
\begin{abstract}
    Large language models (LLMs) may generate outputs that are misaligned with user intent, lack contextual grounding, or exhibit hallucinations during conversation, which compromises the reliability of LLM-based applications. This review aimed to identify and analyze techniques that align LLM responses with conversational goals, ensure grounding, and reduce hallucination and topic drift. We conducted a Rapid Review guided by the PRISMA framework and the PICO strategy to structure the search, filtering, and selection processes. The alignment strategies identified were categorized according to the LLM lifecycle phase in which they operate: inference-time, post-training, and reinforcement learning-based methods. Among these, inference-time approaches emerged as particularly efficient, aligning outputs without retraining while supporting user intent, contextual grounding, and hallucination mitigation. The reviewed techniques provided structured mechanisms for improving the quality and reliability of LLM responses across key alignment objectives.
\end{abstract}

\section{Introduction}

The increasing deployment of LLMs in conversational settings has intensified the need for effective alignment strategies that ensure output consistency with user intent while mitigating hallucinations and topic drift. Hallucinations -- where models generate factually incorrect or fabricated information -- pose significant challenges for real-world applications, especially in domains that demand accuracy and interpretability \citep{huang2025survey,shen2023large,wang2023aligninglargelanguagemodels}. Prior research categorizes hallucinations into types based on factuality and faithfulness, noting that mitigation requires structured interventions, including uncertainty estimation, prompt grounding, and reward-based control \citep{huang2025survey}.

Recent alignment research has investigated methods that operate at distinct stages of the LLM lifecycle: during inference, post-training, and reinforcement learning. 
Inference-time strategies modify generation behavior without updating model parameters, often by applying reward-guided decoding, contrastive scoring, or self-evaluation mechanisms that redirect token selection paths \citep{zhangreinforcement}. 
Post-training methods leverage fine-tuning on curated or synthetic data to guide model behavior toward human preferences while preserving generalization \citep{kumar2025llmposttrainingdeepdive}.
Reinforcement-based methods introduce policy optimization through preference signals or reward models, incorporating optimization techniques \citep{wang2024reinforcement}.

In this paper, we structured as follows. Section~\ref{sec:method} presents the methodological design, including the definition of research questions, the application of the PICO framework, and the procedures for search, filtering, and study selection. Section~\ref{sec:results} describes the findings based on the analysis of selected studies, grouped according to the LLM lifecycle phases. Section~\ref{sec:discussion} interprets these findings by comparing alignment strategies and computational considerations. Finally, the conclusion, Section \ref{sec:conclusion},  summarizes the main contributions and outlines directions for future research.


\section{Methodology}\label{sec:method}
This section presents the methodological foundations that structured this review. The procedures adopted were based on updated guidelines for rapid reviews \citep{garritty2024updated,hamel2021defining}.

Rapid reviews are defined as a form of evidence synthesis that applies tailored methodological restrictions to expedite the review process \citet{garritty2024updated}. 
These reviews typically adopt streamlined methods to address urgent or high-priority questions, incorporating strategies that allow for timely results while explicitly reporting the limitations introduced by such restrictions. 
This review followed the recommendations for rapid review conduct, including the application of the PICO framework to define the scope, adherence to the PRISMA guidelines to ensure transparency in reporting, and the identification of review constraints that were justified and disclosed in accordance with rapid review standards \citep{robinson2011development,selccuk2019guide}. 
These procedures contributed to the systematic selection and interpretation of studies, as outlined in the results section, which grouped findings according to the model pipeline stages and their contribution to the review's guiding questions.

\subsection{Research Question Definition}

The objective of this review was to analyze how alignment techniques in LLMs support conversational objectives, ensure grounded responses, and reduce hallucinations during conversation. 
The purpose is to identify and describe the main approaches proposed in the literature to align LLM outputs with user intentions or topic themes. 
We defined the following questions: 

\begin{itemize}
    \item \textbf{RQ1:} Which techniques or methods align the response of an LLM with the goal or theme of the conversation?
    \item \textbf{RQ2:} What factors support grounded LLM responses related to the conversation topic?
    \item \textbf{RQ3:} How can hallucination or topic deviation from an LLM decrease?
\end{itemize}

\textbf{RQ1} was defined to identify techniques that explicitly aim to adjust LLM behavior to match the goal or theme of a given conversation. 
This includes prompt engineering, decoding strategies, or training-based methods that directly influence how the model maintains relevance and coherence with user intent. 
\textbf{RQ2} was designed to explore which elements contribute to \textit{grounding} the LLM's output in the context provided, such as using external knowledge, structured prompts, or retrieval-augmented inputs. This question targets the mechanisms that improve factual consistency and contextual fidelity. 
\textbf{RQ3} focuses on understanding how hallucinations or topic drift can be reduced through alignment strategies, including filtering mechanisms, preference modeling, or feedback-based optimization. 
Together, these questions guide the review in assessing how current methods contribute to aligning LLM outputs with conversation demands, improving informativeness, and ensuring contextual accuracy.

\subsection{Searching for Papers}
In order to study LLMs alignment in this paper, the PICO framework was adopted, which provided clarity for the development of the search strategy and the selection of studies. 
According to \citet{hamel2021defining}, PICO enables the explicit identification of relevant dimensions of the research question, helping to formulate inclusion criteria in a transparent and reproducible way. Those are the components of PICO:
\begin{itemize}
    \item \textbf{Population}: This component  referred to the entities or subjects of analysis, which needed to be defined clearly to delimit the domain of interest.
    \item \textbf{Intervention}: Identified he specific method, technique, or exposure under investigation.
    \item \textbf{Comparison}: Although optional depending on the review focus, supported contrastive analysis between alternatives or baselines.
    \item \textbf{Outcomes}:  Referred to the expected effects or endpoints assessed in the included studies, forming the central axis of the review findings.
    
\end{itemize}
The objective of this review was to analyze methods and techniques of LLMs alignment that support conversational objectives, ensure grounded responses, and reduce hallucinations. The review sought to identify and describe the main approaches proposed in the literature to align LLM outputs with user intentions. 
According to the objectives and research questions proposed, Table~\ref{tab:pico} presents the PICO stratification based on its components.

\begin{table}[!ht]
\centering
\caption{Research Objective Structured According to the PICO Framework}
\begin{tabular}{|p{3cm}|p{8cm}|}
\hline
\textbf{Component} & \textbf{Definition in this Review} \\
\hline
\textbf{Population (P)} & Papers that aim to introduce and define LLM alignment.\\
\hline
\textbf{Intervention (I)} & RQ1: Identify techniques or methods that align LLM responses with the goal or theme of the conversation; RQ2: Investigate factors that support grounded LLM responses in conversational contexts; RQ3: Identify strategies to mitigate topic deviation and hallucination in LLM outputs. \\
\hline
\textbf{Comparison (C)} & Not applicable.\\
\hline
\textbf{Outcome (O)} &  Identify techniques, methods, or approaches proposed in the literature for aligning LLM outputs.\\
\hline
\end{tabular}
\label{tab:pico}
\end{table}

According to the proposed objectives and research questions, Table~\ref{tab:pico} presents the PICO stratification based on its components. 
This framework supported the definition of the scope of the review and guided the development of the search strategy.
By identifying relevant elements such as the target population, the types of interventions investigated, and the expected outcomes, the PICO structure informed the construction of the search strings used to retrieve studies related to LLM alignment. The formulation of queries sought to ensure the inclusion of papers that proposed or analyzed techniques, methods, or strategies to align LLM responses with conversational goals, support grounding, or reduce hallucinations and topic drift. Table~\ref{tab:query_string} shows the final search strings derived from this formulation.

\begin{table}[!ht]
\centering
\caption{Search Query Used}
\begin{tabular}{|p{0.95\linewidth}|}
\hline
\textbf{Query String} \\
\hline
\parbox[t]{\linewidth}{
("Large Language Model*" OR LLM*) AND alignment
AND\\
(goal AND align*) OR (intention AND align*) OR (response AND align*) OR (response AND tuning) OR (hallucination AND reduction AND (conversation OR response)) OR (mitigating AND hallucination* AND (conversation OR response)) OR context OR (post-training)
NOT\\
value*\\
AND\\
(technique* AND align*) OR (method* AND align*) OR (approach* AND align*)\\
} \\
\hline
\end{tabular}
\label{tab:query_string}
\end{table}


\subsection{Filtering}
The search and filtering process was then structured to ensure that the selected studies matched the scope defined by the PICO components and the formulated research questions.

\begin{figure}
    \centering
    \includegraphics[width=0.8\linewidth]{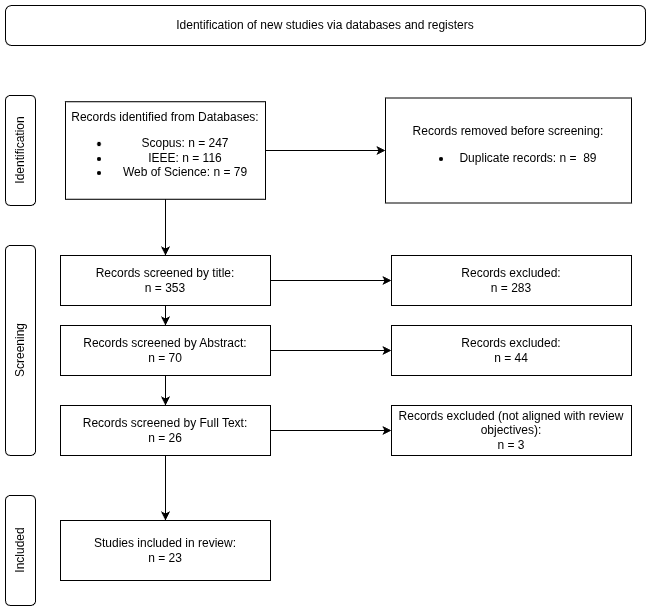}
    \caption{Filtering steps and Database. This image was generated using PRISMA guidelines}
    \label{fig:prisma}
\end{figure}

We searched in three databases: IEEE, Scopus, and Web of Science, as illustrated in Figure~\ref{fig:prisma}, using the query strings shown in Table~\ref{tab:query_string}. The search covered papers published from 2020 to 2025, inclusive. This initial step returned 442 documents, of which 89 were identified as duplicates.

After removing duplicates, we analyzed the titles of the remaining papers. The inclusion criterion for this stage was the presence of relevant keywords from the search query. Based on this criterion, we retained 70 papers for the next phase.

The following step involved reading the abstracts. For this analysis, we used the Parsifal platform, which supported structured evaluation based on the predefined research questions. These were translated into five assessment questions:
\begin{itemize}
    \item Is the main focus of the article to present frameworks, techniques, or methods for aligning the LLM's response with the conversation context or theme?
    \item Does the article focus on alignment approaches that are not jailbreaks, fine-tuning, inference-time methods, moral alignment, multimodal alignment, or question answering?
    \item Does the article address LLM hallucination through a specific form of alignment?
    \item Does the article explore LLM response alignment with the conversation topic, excluding ethical or sentiment-based alignment?
    \item Does the article consider alignment applied to post-training models without involving fine-tuning?
\end{itemize}

Each article received a score depending on how well it addressed these questions: 1.0 point for "Yes" 0.5 for "Partially" and 0.0 for "No". Papers with a total score of 4.5 or 5.0 were selected for the next phase. At this stage, 44 papers were excluded, and 26 were retained.

The final step consisted of a full-text review. In this step, three papers were excluded despite meeting the previous criteria, as their primary objective focused on alignment within specific application areas such as question answering, multimodal systems, or spoken dialogue. After this filtering process, 23 papers remained in the final selection, as presented on Table \ref{tab:final_papers}.

\begin{table}[!ht]
	\centering
	\caption{\textbf{The final list of selected articles.}}
	\label{tab:final_papers}
	\small
	\begin{tabularx}{\textwidth}{|l|X|}
		\hline
		\textbf{Ref.} & \textbf{Short title} \\
		\hline
		\citep{carta2023grounding} & Grounding LLMs in interactive environments with online RL \\
		\citep{munos2023nash} & Nash learning from human feedback \\
		\citep{ahmadian2024back} & Revisiting REINFORCE-style optimization in HF for LLMs \\
		\citep{li2023rain} & RAIN: self-alignment without finetuning \\
		\citep{yuan2023rrhf} & RRHF: Rank responses to align LLMs \\
		\citep{linunlocking} & Unlocking base LLMs via in-context learning \\
		\citep{khanov2024args} & ARGS: Alignment as reward-guided search \\
		\citep{lin2024mitigating} & Mitigating the alignment tax of RLHF \\
		\citep{zhang2024self} & Self-alignment for factuality using self-evaluation \\
		\citep{khaki2024rs} & RS-DPO: Rejection sampling + DPO for alignment \\
		\citep{guo2023beyond} & Beyond imitation: fine-grained quality signals for LLM alignment \\
		\citep{sun2024salmon} & SALMON: Self-alignment with instructable reward models \\
		\citep{zhang2024can} & Can LLM graph reasoning generalize beyond memorization? \\
		\citep{wang2024codeclm} & CodecLM: Alignment with synthetic tailored data \\
		\citep{liu2024direct} & Contrastive prompt distillation with self-rewarding \\
		\citep{gao2024linear} & Linear alignment: closed-form alignment without tuning \\
		\citep{dong2024unsupervised} & Unsupervised LLM alignment via contrastive feedback \\
		\citep{wang2024hybrid} & Hybrid alignment training for LLMs \\
		\citep{zhengimproving} & Group-invariant learning to improve alignment generalization \\
		\citep{xu2024automatic} & Automatic pair construction for contrastive post-training \\
		\citep{naseem2024grounded} & Grounded preference model for LLM alignment \\
		\citep{pang2024arm} & ARM: Alignment with residual energy-based models \\
		\citep{wong2024aligning} & Bayesian inference for aligning human feedback in code generation \\
		\hline
	\end{tabularx}
\end{table}

\section{Results}\label{sec:results}
This section presents the results obtained through the analysis of the 23 studies selected according to the criteria described in Section~\ref{sec:method}. Each paper was individually examined with a focus on how it contributed to answering the three research questions formulated in this review. The analysis prioritized identifying whether the proposed methods aimed to align the model's behavior with the conversation’s objective \textbf{RQ1}, supported grounding of the outputs in the prompt or contextual information \textbf{RQ2}, or reduced hallucinations and topic drift \textbf{RQ3}.
To improve interpretability, the papers were grouped into subsections based on the stage of the language model pipeline in which alignment was applied: inference-time, post-training, or reinforcement learning.

Each alignment phase addressed different aspects of model behavior and required distinct mechanisms. Inference-time methods generally prioritized prompt structuring and decoding strategies to guide the model without parameter updates, while post-training approaches applied fine-tuning on curated or synthetic datasets to induce preference-consistent behaviors. Reinforcement learning methods incorporated reward-guided policy optimization to iteratively adapt model outputs based on feedback signals. Among these, reinforcement learning required the most computational resources due to the presence of multiple models, online sampling procedures, and iterative optimization steps involving gradient updates and policy evaluation.

\subsection{Inference-Time Methods}

\subsubsection{Untuned LLMs with Restyled In-context ALign-
ment (URIAL)}

\citet{linunlocking} proposed \texttt{uril} as a tuning-free alignment method that relied exclusively on in-context learning with a static prompt structure. 
The approach employed a \textit{system prompt} followed by a fixed set of stylistically crafted examples to guide the base LLM's behavior, without using \texttt{sft} or \texttt{rlhf}. 
The system prompt, adapted from Llama-2-chat, defined the assistant’s role and behavioral norms, emphasizing helpfulness, respect, honesty, safety, and the refusal to engage in harmful or controversial content. It also required responses to be well-structured, informative, and socially responsible. This fixed instruction was added to the prompt as a stable context preceding all example interactions.

The in-context examples were carefully constructed to replicate stylistic patterns observed in aligned LLMs \citep{linunlocking}. Each example began by affirming the user's input and introducing background content, followed by an organized body using enumeration formats (e.g., \texttt{[n]. [topic]: [details]}), and concluded with a summary embedding safety disclaimers. These demonstrations incorporated stylistic tokens frequently shifted by alignment methods, such as “Hello,” “Thank,” “However,” and “Remember,” to encourage polite and safe discourse. Examples also illustrated how to respond to morally sensitive queries, offering empathetic language, constructive redirection, and safety-oriented reasoning rather than direct refusals. The authors demonstrated that using as few as $K=1$ to $K=3$ examples yielded effective alignment, with $K=3$ resulting in approximately 1{,}011 tokens for the static prefix. 
The fixed nature of the prompt allowed for caching, improving efficiency compared to retrieval-based in-context learning. 
Additionally, the framework supported multi-turn conversations by appending prior history as new examples, without changing the base prefix.

\texttt{uril} \citep{linunlocking} addressed \textbf{RQ1} by aligning model responses to user intent and thematic structure through in-context stylistic imitation. For \textbf{RQ2}, the method supported grounded outputs by explicitly encoding expected response formats and behavior constraints in the system prompt and examples. In relation to \textbf{RQ3}, hallucinations and topic deviations were mitigated by consistently conditioning the model with safe and structured exemplars. Unlike methods that relied on parameter updates, \texttt{uril} achieved comparable or superior performance to \texttt{sft}- and \texttt{rlhf}-aligned models, particularly when applied to well-pretrained base models such as Mistral-7b and Llama-2-70b, reinforcing the viability of prompt-based alignment for conversation fidelity.

\subsubsection{RAIN}

\texttt{rain}, introduced by \citet{li2023rain}, pursued alignment without finetuning by implementing a self-evaluation and rewind mechanism at inference. During generation, the model assessed the alignment of its output with human-defined goals and selectively rewrote parts that received low scores. This inference-time procedure used heuristic simulations and backward rewrites to revise token paths and increase adherence to alignment objectives such as safety and factuality. Instead of guiding behavior statically as in \texttt{uril}, \texttt{rain} dynamically corrected generations through token-level self-assessment. This process relied on structured prompts that communicated human goals and enabled the model to score its own outputs, thus integrating evaluation directly into the token selection process. These prompts, although not designed via a formal methodology, were provided as fixed templates and included label-swapping techniques to reduce evaluation bias.

Despite not requiring parameter updates or external datasets, \texttt{rain} \citep{li2023rain} presented computational limitations. The method increased inference time due to its inner loop, which queried the model multiple times for candidate tokens and conducted self-evaluation for each path. On models like LLaMA 30B, the approach required roughly four times more processing time than standard inference. Although this overhead diminished with safer models, it remained a trade-off for runtime alignment. Unlike memory-intensive alignment methods such as \texttt{rlhf}, \texttt{rain} preserved memory efficiency by avoiding gradient tracking and auxiliary model components. Nevertheless, the authors acknowledged that this time complexity might constrain deployment and suggested that future work could use \texttt{rain} to generate alignment data for downstream fine-tuning, thereby offloading the computational cost from inference to training.

\texttt{rain} \citep{li2023rain} contributed to \textbf{RQ1} by integrating alignment directly into the token selection process through structured self-evaluation during inference. For \textbf{RQ2}, the method grounded generation by using prompts that encoded human-aligned evaluation criteria such as harmlessness and factual accuracy. In addressing \textbf{RQ3}, \texttt{rain} reduced hallucinations and misalignment by selectively rejecting and regenerating low-scoring continuations through its rewind mechanism. By enabling frozen models to self-correct without access to external alignment data or finetuning, \texttt{rain} demonstrated a viable path for runtime alignment in resource-constrained or high-stakes inference scenarios.

\subsubsection{ARGS}

Similarly to \texttt{rain}, the \texttt{args} method proposed by \citet{khanov2024args} avoided reinforcement learning by integrating alignment directly into the decoding process through a reward-guided scoring function. At each generation step, ARGS modified the base LLM's token probabilities by combining the log-likelihood with a scalar reward from an external reward model. This reward reflected human preferences, and the combined score determined the next token during inference. The reward model was trained separately using paired comparisons from datasets such as HH-RLHF and SHP, with annotations indicating which response better aligned with helpfulness and harmlessness criteria. Unlike training-based methods like \texttt{rlhf}, ARGS avoided gradient updates, fine-tuning, or auxiliary models, and introduced alignment post hoc during inference without changing the language model’s parameters.

The authors acknowledged that ARGS \citep{khanov2024args} exhibited limitations related to both scope and implementation. The method was evaluated primarily on standard alignment benchmarks, and its generalization to more complex, multi-step reasoning tasks remained untested. The framework also depended on the quality of the external reward model; any inaccuracy in this component directly influenced the alignment of generated outputs. Although ARGS avoided expensive training costs, it introduced increased computational complexity during decoding. The integration of reward-guided scoring required evaluating a top-$k$ set of candidate tokens at each step, leading to higher inference times compared to classical decoding. For example, using OPT-2.7b with $k=10$ led to a 1.9-fold increase in generation time relative to standard greedy decoding. While this cost remained tractable, it represented a computational trade-off in achieving decoding-time alignment.

\texttt{args} \citep{khanov2024args} contributed to \textbf{RQ1} by incorporating alignment signals into the decoding algorithm, adjusting token selection through a reward-guided function. It supported \textbf{RQ2} by preserving coherence via language model likelihood while incorporating preference-based criteria from the reward model, maintaining thematic relevance. For \textbf{RQ3}, the calibrated scoring function and the flexibility of the $w$ parameter allowed the model to suppress low-reward continuations, which reduced hallucinations and topic drift. By enabling post hoc alignment with trained reward models and without modifying model parameters, ARGS offered a decoding-time alternative to alignment that complemented approaches like \texttt{uril} and \texttt{rain}.

\subsubsection{Linear Alignment}

\citet{gao2024linear} presented Linear Alignment, a method that applied self-contrastive decoding to shift the output distribution of a LLM in a direction consistent with preference principles. This was achieved without training or reward model supervision. The method estimated the alignment adjustment through two forward passes per token—one with a principle prompt and one without—to compute a closed-form update to the token logits. The resulting direction, constrained by a divergence boundary, altered the output distribution while preserving semantic coherence. This procedure leveraged the model’s understanding of the principle prompt to extract a value gradient and align generation accordingly. The approach aligned with earlier inference-time alignment methods such as \texttt{args} and \texttt{rain}, but differed by producing a direct optimization update without external evaluation components.

Although Linear Alignment \citep{gao2024linear} eliminated the need for data annotation and fine-tuning, its effectiveness depended on the \texttt{sft} model's capacity to interpret principle prompts. The alignment process incurred a computational cost, approximately doubling the inference time due to the dual forward passes required for each token. While the authors noted that inference could be parallelized, this adjustment introduced additional GPU usage. The method also showed sensitivity to the step-size hyperparameter that controlled the update magnitude. Miscalibration could either weaken alignment or induce overcorrection, although the method exhibited general stability once tuned. The authors further reported mixed effects when applying Linear Alignment to pre-aligned models such as Llama2-13B-Chat. In some personalized settings, alignment performance declined, likely due to prior optimization toward generalized helpfulness and harmlessness objectives that constrained the model's adaptability to individual preferences. Additionally, in perplexity-based tasks, the linear adjustment of token probabilities occasionally produced inaccurate outputs when the target tokens had uniformly low base probabilities.

Linear Alignment \citep{gao2024linear} responded to \textbf{RQ1} by directly modifying the generation policy based on preference prompts without modifying model parameters. For \textbf{RQ2}, it incorporated contextual and principle information into a bounded update mechanism that maintained the semantic intent of the original output while aligning to preference-based objectives. In relation to \textbf{RQ3}, it constrained hallucinations and misalignment by limiting the output distribution shift through divergence control, while avoiding reinforcement learning. The method contributed to the class of inference-time alignment strategies by demonstrating that preference optimization can be approximated in a closed form, leveraging prompt-based cues without reliance on external feedback or training data.

\subsection{Post-Training Methods}

\subsubsection{DLMA}

\citet{liu2024direct} proposed \texttt{dlma} as a fine-tuning-based alignment method that combined contrastive prompt pairs, a self-rewarding scoring mechanism, and a revised \texttt{dpo} loss function. The method applied a two-phase training pipeline: first, supervised fine-tuning (SFT) on a downstream instruction dataset produced a base model
$\pi_{\text{SFT}}$ 
with general instruction-following capabilities. 
Second, DLMA generated paired responses from contrastive prompts—one aligned with target behaviors such as helpfulness or harmlessness, and one deviating from them. The model then computed a self-rewarding score based on generation probabilities to quantify preference between the responses. This preference signal was integrated into a modified DPO loss to optimize the model toward preferred outputs without requiring external human labels or separate reward models. By using only model-generated data and internal scoring, DLMA operationalized alignment through self-supervision during fine-tuning.

While DLMA \citep{liu2024direct} avoided the complexity of \texttt{rlhf} pipelines, it introduced certain limitations. 
Its alignment performance was validated on LLaMA2-7B and LLaMA2-13B, but scalability to larger models remained untested. 
The self-rewarding score, which was effective for model-generated responses, did not perform well on externally sourced text, constraining its generalization. 
Theoretical assumptions supporting the self-rewarding formulation were tailored to model outputs and might not hold under broader conditions. 
Additionally, improvements diminished across iterative alignment rounds, suggesting convergence to a plateau in gain. Although DLMA reduced reliance on labeled datasets, it still required considerable compute resources: training a 7B-parameter model completed in approximately eight hours using eight A100 80G GPUs. These constraints positioned DLMA as more efficient than \texttt{rlhf} but still dependent on infrastructure unavailable in low-resource environments.

\texttt{dlma} \citep{liu2024direct} contributed to \textbf{RQ1} by aligning pretrained LLMs through fine-tuning with preference signals derived from contrastive prompts and internal evaluation. 
It supported \textbf{RQ2} by reinforcing grounded behavior based on prior instruction tuning, using prompt structures that emphasized target attributes. 
For \textbf{RQ3}, the method reduced hallucinations by prioritizing self-evaluated helpful outputs and suppressing unsafe or irrelevant completions. 
DLMA exemplified an alignment strategy that leveraged internal model behavior for optimization, standing between inference-only alignment techniques and full supervision pipelines.

\subsubsection{SAF}

\citet{zhang2024self} proposed Self-Alignment for Factuality as a fine-tuning method that aligned LLMs by leveraging internal evaluation capabilities to generate preference data without requiring human-annotated factuality labels. The framework operated by sampling multiple candidate responses for each prompt, then using a self-evaluation module, termed SELF-EVAL, to assess the truthfulness of each output through internally calibrated True/False prompts. 
To improve confidence estimation and calibration, the method introduced Self-Knowledge Tuning (SK-TUNING), which fine-tuned the model on factuality evaluation data automatically derived from heterogeneous sources such as Wikipedia and BIG-bench. The resulting confidence scores were used to rank candidate responses, forming preference pairs for training with the Direct Preference Optimization (DPO) algorithm. 
This closed the alignment loop by grounding the model in its own self-assessed outputs and allowing alignment with factual behavior based solely on model-internal signals.

The method \citep{zhang2024self} demonstrated alignment improvements without relying on external supervision, but several limitations were acknowledged. The validation focused on 7B-parameter models from the LLaMA family, and its effectiveness on larger or already aligned models, such as LLaMA2-Chat, remained unverified. The self-evaluation module, although enhanced by SK-TUNING, still produced factual errors under specific conditions. Manual error analysis revealed persistent issues such as failures in disambiguating misleading premises, providing confident answers to ambiguous or debatable questions, and reproducing superstitions or cultural biases. These errors likely reflected limitations in pretraining data coverage and indicated the need for complementary strategies, such as instructing the model to express uncertainty or deferral. Additionally, while SK-TUNING improved calibration, the authors noted that further research into more efficient confidence estimation mechanisms could advance the framework. The method also left open the possibility of integrating decoding-time approaches for better factual consistency.

This self-alignment approach \citep{zhang2024self} addressed \textbf{RQ1} by implementing fine-tuning based on internal model evaluations rather than external supervision. For \textbf{RQ2}, it reinforced grounded behavior by using SK-TUNING to calibrate the model’s confidence in its factual responses. In addressing \textbf{RQ3}, the framework reduced hallucinations through preference optimization that favored high-confidence completions. By operationalizing factual self-evaluation as a training signal, the method demonstrated an alternative alignment pathway rooted in model introspection rather than externally imposed criteria.

\subsubsection{RS-DPO}

\citet{khaki2024rs} introduced RS-DPO as a training-time alignment method that integrated rejection sampling with \texttt{dpo} to refine the outputs of a supervised fine-tuned LLM. 
The method first applied \texttt{sft} on instruction-following data to establish a base model. 
Then, for each prompt, it generated multiple candidate responses, scored them using a trained reward model, and selected preference pairs based on reward gaps exceeding a defined threshold. 
These synthetic preference pairs, constructed from model-generated completions, were used to train the policy model using DPO. 
This allowed RS-DPO to align the model with high-quality completions by directly optimizing the policy for the preferred responses. Unlike DPO variants dependent on manually annotated or externally sourced response pairs, RS-DPO generated its alignment data internally, relying solely on a reward model to supervise the \texttt{rs} stage.

Despite its training efficiency and stability relative to \texttt{rlhf}, RS-DPO \citep{khaki2024rs} presented limitations in scope and scale. The method was evaluated primarily on helpfulness objectives using open-source preference datasets, and its ability to generalize to other alignment dimensions, such as harmlessness, remained untested. Additionally, all experiments focused on 7B-parameter models, and the framework had not been extended to larger or proprietary LLMs. 
The reliance on existing human-annotated prompt-response datasets for both reward model training and preference data generation constrained the framework’s capacity to operate in domains lacking well-curated data. Although the reward model used in RS-DPO contributed to stable sampling of high-quality preference pairs, the method still depended on maintaining sampling within the reward model’s distribution, which could limit flexibility in unfamiliar prompt domains.

RS-DPO \citep{khaki2024rs} addressed \textbf{RQ1} by implementing alignment through fine-tuning with reward-scored preference pairs generated via rejection sampling. It contributed to \textbf{RQ2} by enforcing human-preference consistency over multi-turn prompts present in datasets like OASST1 and HH-RLHF. Regarding \textbf{RQ3}, the use of a trained reward model to select completions with high relative quality reduced the likelihood of hallucinations by optimizing for factual and relevant content. The method offered a compromise between data efficiency and behavioral alignment, positioning itself between traditional RLHF and DPO variants that require handcrafted or model-agnostic preference data.

\subsubsection{ARM}

\citet{pang2024arm} proposed the \texttt{arm} as a post-training method to align LLMs without relying on reinforcement learning. ARM constructed a residual energy-based target distribution by combining the output probabilities from a supervised fine-tuned policy with a reward-based energy term. This formulation allowed the authors to sample expert-like completions through self-normalizing importance sampling, favoring high-reward outputs. The method then trained a new policy to imitate these samples using preference-based objectives such as \texttt{dpo}. The reward model guided both the construction of the target distribution and the preference scoring of sampled outputs. Unlike reinforcement-based approaches, ARM achieved alignment through fully offline transformations, relying only on static preference data and reward-guided sampling.

 ARM's \citep{pang2024arm} performance varied depending on the preference modeling strategy, with the Bradley-Terry formulation outperforming Plackett-Luce. This gap was attributed to the increased sensitivity of Plackett-Luce to noise in reward model outputs, especially when handling multi-way comparisons. Although ARM aimed to improve controllability and alignment, the resulting models retained the possibility of producing harmful content. The authors acknowledged that further safety evaluations were necessary before real-world deployment. While ARM showed flexibility by accommodating low-resource scenarios and simplifying the alignment pipeline, it still depended on reward model quality and curated prompt datasets to maintain alignment fidelity.

ARM \citep{pang2024arm} contributed to \textbf{RQ1} by providing a direct alignment mechanism through energy-based sampling and preference-based fine-tuning. It addressed \textbf{RQ2} indirectly by modeling alignment through reward signals derived from human preferences, rather than explicit conversation grounding. For \textbf{RQ3}, ARM mitigated hallucinations by selecting high-reward completions, although its effectiveness depended on how well the reward model penalized undesired responses.

\subsubsection{Constrative}

\citet{xu2024automatic} proposed a contrastive post-training strategy that aligned LLMs without relying on reward models or human-labeled preferences. The method automatically generated preference pairs by sampling outputs from LLMs of different capabilities, assuming responses from stronger models were preferable. Using \texttt{dpo}, the authors fine-tuned models to increase the relative probability of preferred outputs over less preferred ones. The training process incorporated a curriculum that gradually introduced harder comparisons, starting with pairs where the quality gap was wider. This approach allowed the model to progressively learn finer distinctions. The authors avoided reinforcement learning due to issues such as reward hacking and poor generalization, which they encountered during experiments with \texttt{rlaif}. DPO, in contrast, required no intermediate reward modeling and operated directly on pre-constructed contrastive data.

The fine-tuning stage followed initial supervised instruction tuning and involved models such as LLaMA-7B and Orca-13B \citep{xu2024automatic}. In both small- and large-scale experiments, DPO yielded significant performance improvements over supervised baselines. The authors observed that initializing DPO from a supervised model further enhanced outcomes. To manage computational load, they employed DeepSpeed ZeRO-3 and distributed training over multiple GPUs. Although the method involved high-end infrastructure, it remained more computationally efficient than \texttt{rlhf} by avoiding online sampling and reward model training. The curriculum learning strategy supported stable convergence, and the authors noted that performance gains persisted even after supervised fine-tuning had saturated.

This method \citep{xu2024automatic} addressed \textbf{RQ1} by directly optimizing preference-consistent behavior through contrastive fine-tuning without requiring human labels. It supported \textbf{RQ2} by grounding learning in outputs from stronger models, assuming these responses reflected human-aligned behavior. For \textbf{RQ3}, the model reduced hallucinations by increasing the likelihood of higher-quality completions, selected through inter-model comparisons. The absence of prompt design as an alignment mechanism emphasized a training-based strategy that complemented methods relying on structured prompting or in-context examples.

\subsubsection{RRHF}

\citet{yuan2023rrhf} proposed \texttt{rrhf} as a method to align LLMs with human preferences using a ranking-based loss over sampled responses. The approach extended supervised fine-tuning by incorporating a ranking loss that compared conditional log probabilities across multiple candidate completions, encouraging the model to assign higher likelihood to responses with better human preference scores. During training, RRHF received as input a set of responses per query, scored these responses using the model’s own length-normalized conditional log probabilities, and aligned the score rankings with external human preference or proxy reward labels. This alignment objective removed the need for a reward model or reinforcement learning algorithm, simplifying the training procedure. 
The method also proved effective in offline learning settings, as shown by the Wombat model, which was trained on responses from ChatGPT and evaluated on Alpaca prompts.

The method \citep{yuan2023rrhf} demonstrated dual functionality. It served as a generation model aligned with preferences and also performed as a reward model by assigning scores to responses that correlated with existing human-preference annotations. By avoiding policy gradient steps and multiple model components required in \texttt{ppo}, RRHF reduced memory and computation demands. However, it required multiple response candidates per query, increasing per-query GPU usage. The model was trained offline in 4 to 6 hours using 8 A100 GPUs, and the authors reported comparable or better performance than PPO in human and automatic evaluations. Although online variants of RRHF were also explored, they introduced additional complexity and resource costs that countered the method’s initial motivation. The use of pre-existing prompts and the design of an evaluation prompt for ChatGPT-based reward scoring illustrated that prompt design was limited to the evaluation stage rather than being central to the alignment method itself.

This method \citep{yuan2023rrhf} addressed \textbf{RQ1} by replacing policy optimization with a ranking-based fine-tuning approach that increased the probability of preferred outputs. It contributed to \textbf{RQ2} by incorporating reward-aligned examples and leveraging model-based scoring for quality assessment. For \textbf{RQ3}, it reduced hallucinations by penalizing low-preference responses during training. RRHF aligned with other fine-tuning-based approaches, diverging from contrastive methods by using rank matching rather than DPO-style pairwise optimization, and offering a scalable strategy with reduced architectural overhead.

\subsubsection{FIGA}

The \texttt{figa} method, proposed by \citet{guo2023beyond}, introduced alignment through token-level supervision based on differences between initial and revised responses. The approach contrasted low-quality outputs with minimally revised high-quality versions generated using curated prompts and a stronger LLM. Using Levenshtein distance, the method identified added, deleted, and substituted tokens, associating them with positive or negative alignment signals. A loss function then incorporated these fine-grained rewards and penalties to adjust token probabilities during fine-tuning. This procedure enabled the model to learn more precise behavioral corrections than traditional response-level losses and was implemented using a dataset—SPA—constructed via prompt-guided revision and filtered using a static reward model. The reward model did not participate in optimization but ensured that selected training samples showed measurable quality gaps. 

Prompt design was required to instruct the stronger model to revise low-quality completions minimally, based on predefined issues such as lack of detail or factual errors. Each issue had a corresponding prompt designed to elicit targeted improvements, and these revisions served as supervision for fine-tuning. While the reward model did not enter training as an optimization signal, its use in filtering pairs reinforced that FIGA \citep{guo2023beyond} employed reward-guided data construction rather than direct optimization. Compared to \texttt{ppo} and other \texttt{rlhf} methods, FIGA reduced training complexity by removing the need for rollout and value models. It achieved competitive performance using supervised learning alone, with experimental results indicating lower computational costs and improved stability. The construction of the SPA dataset incurred minimal financial cost, and the training setup relied on a single model with a batch size of 128, avoiding the multi-model setup typical in \texttt{rlhf}.

\texttt{figa} \citep{guo2023beyond} contributed to \textbf{RQ1} by aligning behavior through token-specific feedback obtained from expert-guided revision, rather than full-sample preference comparison. It addressed \textbf{RQ2} by reinforcing grounded and contextually appropriate content using data filtered by a reward model. For \textbf{RQ3}, the token-level penalties imposed during fine-tuning suppressed hallucinated or off-topic content directly. The method differed from ranking- or preference-based fine-tuning approaches such as \texttt{rrhf} and DPO by applying alignment at a finer granularity, enabling more precise corrections while remaining computationally efficient.

\subsubsection{CodeCLM}

\citet{wang2024codeclm} proposed CodecLM as a framework for alignment via synthetic instruction tuning, using a strong LLM to generate high-quality instruction-response pairs without human annotation. The method encoded seed instructions into metadata, representing task-specific features such as use case and required skills. These metadata guided the generation of tailored instructions by prompting the strong model. Prompt design was central to the framework: separate templates supported the stages of encoding metadata, generating instructions, constructing rubrics for complexity, and improving instruction quality. To refine the data quality, the authors implemented Self-Rubrics to enhance instruction complexity and Contrastive Filtering to prioritize examples with a measurable performance gap between the target and strong models. The final synthetic dataset, derived through this multistage process, was used to fine-tune the target LLM, resulting in consistent gains across open-domain instruction-following benchmarks.

CodecLM \citep{wang2024codeclm} operated as a batch fine-tuning approach rather than as a real-time adaptive system. However, the authors acknowledged distribution mismatch challenges in live settings and suggested a strategy for continual alignment: by collecting user instructions or feedback during deployment and feeding them back into the CodecLM pipeline, it would be possible to regenerate tailored data for further fine-tuning. This approach offered a mechanism for periodic adaptation without requiring on-the-fly updates. In contrast to \texttt{rlhf}-based alignment pipelines that require multi-model setups and online sampling, CodecLM emphasized data quality and alignment via fine-tuning on contrastively selected examples. The authors demonstrated lower computational overhead, reporting training setups with eight A100 GPUs and low data generation cost, and positioned the method as a generalizable, domain-adaptable strategy for aligning models toward specific conversational goals.

This method \citep{wang2024codeclm} contributed to \textbf{RQ1} by fine-tuning LLMs on instruction-response pairs specifically tailored to downstream tasks, enabling alignment to user goals. For \textbf{RQ2}, the metadata-guided generation process grounded the responses in contextual relevance, while contrastive filtering ensured domain-specific weaknesses were addressed. Regarding \textbf{RQ3}, the systematic curation and filtering of instructions helped remove content likely to cause hallucination by refining the target model's attention to known limitations. CodecLM differed from preference-optimization methods by shifting alignment toward synthetic data engineering and offline optimization, and it complemented token-level or ranking-based methods by demonstrating that prompt-driven data curation alone could yield effective post-training alignment.

\subsection{Reinforcement Learning Based methods}
\subsubsection{REINFORCE}

In the context of \texttt{rlhf}, \citet{ahmadian2024back} challenged the dominance of \texttt{ppo} by introducing REINFORCE and its variant \texttt{rloo} as efficient alternatives for LLM alignment. These methods aligned model behavior through policy gradient optimization, where entire sequences were treated as single actions and optimized using reward signals from preference models. RLOO improved stability by applying a leave-one-out strategy to baseline estimation. Both approaches addressed \textbf{RQ1} by fine-tuning LLMs to prefer high-reward sequences without requiring critic networks or complex PPO configurations. Their use of KL penalties constrained divergence from the initial supervised policy, thereby preserving prompt relevance and addressing \textbf{RQ2}. They also mitigated hallucination and topic deviation for \textbf{RQ3} through controlled updates and variance reduction. This contrasted with more structured preference modeling seen in \citet{munos2023nash}, where alignment emerged from policy convergence to Nash equilibrium within a preference framework.

\citet{munos2023nash} proposed the \texttt{nlhf} framework as an alternative to traditional RLHF approaches, aiming to align large language model policies by directly optimizing for the Nash equilibrium of a learned preference model. 
The method used training-time optimization, relying on preference comparisons between pairs of responses to learn a policy that consistently generated responses preferred over alternatives. 
The authors introduced gradient-based algorithms, including Nash-MD-PG and Nash-EMA-PG, which updated the LLM policy using preference model gradients regularized by KL-divergence from a reference policy. 
This alignment strategy addressed \textbf{RQ1} by guiding policy optimization toward equilibrium strategies that outperformed competing policies in terms of human preference. 
It supported \textbf{RQ2} by using context-conditioned preference evaluations that preserved relevance to the input prompts and maintained proximity to the supervised fine-tuned baseline through KL-regularization. 
The approach addressed \textbf{RQ3} by reducing hallucinations and topic drift, as the preference model penalized off-topic completions and KL penalties limited deviation from grounded initial policies. Experiments conducted on summarization tasks demonstrated consistent alignment with human preferences and improvements in response coherence, suggesting that NLHF provided an effective and stable alternative to reward-based RLHF pipelines.

\subsubsection{GPM}

\citet{naseem2024grounded} extended RLHF alignment by introducing a \texttt{gpm} trained on synthetic entailment-based data to reward responses that adhered to retrieved evidence. The GPM served as the reward signal in a PPO loop and fine-tuned a model to improve contextual relevance. This method addressed \textbf{RQ1} by applying RLHF with a custom reward function optimized for grounding, \textbf{RQ2} by ensuring that rewarded responses aligned with retrieval context, and \textbf{RQ3} by discouraging outputs unverified by the supporting evidence. This shared similarities with the principle-based reward strategy in \citet{sun2024salmon}, where the SALMON framework trained a reward model to reflect explicit human principles and guided alignment through reinforcement learning. While both approaches avoided traditional human annotation pipelines, SALMON generalized alignment objectives through rule-based formulations, whereas the GPM focused on retrieval-specific grounding.

\citet{sun2024salmon} proposed the \texttt{salmon} framework as a training-time alignment method that replaced human preference annotations with principle-driven feedback to guide reinforcement learning. The method involved training a reward model to assign scores based on explicit human-defined principles and using these scores to optimize a policy model. During the alignment process, the policy generated responses which were evaluated against dynamic principles, and the reward model provided feedback that shaped future updates. This approach addressed \textbf{RQ1} by aligning model behavior with intended goals through a reward structure defined by normative principles rather than individual preference data. It supported \textbf{RQ2} by encoding principles that emphasized relevance, task focus, and contextual accuracy, thereby grounding outputs according to pre-specified criteria. \textbf{RQ3} was addressed through the introduction of intervention principles that penalized undesired behaviors such as hallucination, speculation, or digression, thus guiding the model toward more controlled and reliable outputs. While SALMON offered scalability and removed dependence on proprietary annotated data, it introduced complexity in principle design and raised challenges in managing context sensitivity and reliability without external factual verification.

\subsubsection{RLCF}

\citet{dong2024unsupervised} introduced the \texttt{rlcf} framework as a post-pretraining alignment method tailored for Information Retrieval tasks. This method aligned LLM responses by grouping similar documents and applying reinforcement learning based on contrastive signals constructed from their internal differences. The training relied on a group-wise reciprocal rank reward function, which evaluated the model’s capacity to generate outputs that accurately reflected distinctions within each group. RLCF addressed \textbf{RQ1} by guiding model optimization with task-specific objectives without requiring labeled data, aligning outputs to reflect content differences among similar documents. It addressed \textbf{RQ2} by encouraging groundedness through contrastive feedback, which pushed the model to generate document-specific responses rather than generalizations. \textbf{RQ3} was addressed by reducing hallucinations and topic drift, since the model received reinforcement only when its outputs captured unique aspects of each input. This method demonstrated that contrastive ranking signals could serve as effective unsupervised supervision for aligning LLMs in domains where inputs have high similarity but require nuanced differentiation.

\citet{wong2024aligning} advanced alignment in code generation tasks through the \texttt{cRLHF} framework, which replaced single-evaluator feedback with crowd-sourced input and Bayesian aggregation to compute correctness-based rewards. This reward signal guided \texttt{ppo} fine-tuning and addressed \textbf{RQ1} by aligning model output through line-level supervision. \textbf{RQ2} was addressed by reinforcing output consistency with task specifications, and \textbf{RQ3} was mitigated by penalizing error-prone generations. This crowd-sourced supervision paralleled the unsupervised strategy adopted in \citet{dong2024unsupervised}, where \texttt{rlcf} used contrastive signals from similar document clusters to define task-specific alignment objectives. Both strategies expanded feedback mechanisms beyond traditional reward models but diverged in application, with RLCF focusing on IR-specific ranking and cRLHF targeting syntactic correctness in code.

\citet{carta2023grounding} introduced GLAM, a method for aligning LLMs through functional grounding in interactive environments. They used pretrained FLAN-T5 models as policies and refined them using PPO based on task rewards. This method addressed \textbf{RQ1} by iteratively aligning responses through interaction with structured tasks. It supported \textbf{RQ2} by encoding domain structure and goals into the input prompts and using environmental dynamics to teach conceptual dependencies. GLAM addressed \textbf{RQ3} by reducing reliance on static datasets and mitigating the emergence of spurious correlations, thereby suppressing ungrounded or off-topic actions. The contrast between PPO-driven imitation (Behavioral Cloning) and reward-guided exploration highlighted the benefits of functional grounding through reinforcement learning in textual environments.

\subsubsection{HMA}

\citet{lin2024mitigating} introduced \texttt{hma}, a post-training alignment method that merged pre-trained and RLHF-tuned weights using part-specific interpolation. HMA aligned responses \textbf{RQ1} by averaging model layers instead of retraining, preserving base model capabilities while reinforcing alignment. It supported groundedness \textbf{RQ2} by restoring factual reasoning features through selective weight averaging and reduced hallucination \textbf{RQ3} by balancing alignment-specific adaptation and retained general skills. This method addressed the alignment-forgetting trade-off more efficiently than traditional RLHF or DPO approaches. Similar in post-training scope, \citet{zhang2024can} applied DPO to graph reasoning tasks using labeled correct-incorrect response pairs to fine-tune models after instruction tuning. While both methods operated post-training, DPO updated model preferences based on correctness, whereas HMA restored pre-trained abilities lost during alignment.

\citet{zhang2024can} applied \texttt{dpo} as a post-training alignment method to improve large language models' performance on graph reasoning tasks. This approach fine-tuned models that had already undergone instruction tuning, using labeled input-response pairs where one answer was correct and the other incorrect. The training objective favored responses labeled as correct, aligning the model’s preferences with predefined correctness criteria. DPO addressed \textbf{RQ1} by adjusting the model’s output probabilities to favor accurate responses over incorrect ones, thereby reinforcing alignment with task-specific goals. It supported \textbf{RQ2} by grounding model outputs in structured graph data, training the model to respond based on semantic, numerical, or structural correctness as defined by the labeled examples. Regarding \textbf{RQ3}, DPO reduced hallucinated or erroneous completions within the scope of structured reasoning problems by providing clear correctness-based feedback, though it did not address hallucinations in open-domain generation. The method improved generalization from synthetic to real-world graph data but faced limitations in consistency across diverse evaluation settings.

\citet{wong2024aligning} and \citet{sun2024salmon} both implemented policy learning based on structured feedback, but their alignment objectives varied. \texttt{cRLHF} focused on correctness using crowd judgments, while SALMON relied on designer-specified principles and used synthetic labels. In both, reinforcement learning replaced fixed reward models, introducing adaptable and scalable alignment systems. These approaches shared with \citet{zhengimproving} a concern with variance control and generalization. In their method, group-invariant learning and adaptive KL penalties regularized updates during RLHF, aligning model performance across diverse data subsets \textbf{RQ1}, preserving relevance through KL-regularization \textbf{RQ2}, and addressing hallucination via constraint-based exploration \textbf{RQ3}. Together, these strategies formed a class of alignment techniques prioritizing flexibility, generalization, and robustness in policy optimization.

\citet{lin2024mitigating} and \citet{wang2024hybrid} focused on integrating supervised and preference-based training objectives. The \texttt{hbat} method alternated instruction-following and preference-based fine-tuning using Elastic Weight Consolidation to retain knowledge across training rounds. It aligned responses \textbf{RQ1} through structured training stages, supported grounding \textbf{RQ2} by combining instruction fidelity with human preference adherence, and reduced hallucination \textbf{RQ3} by balancing updates and preventing overfitting to either training objective. This integrated optimization echoed the trade-off-aware approach seen in \texttt{hma} and extended it to full training loops. Across all studies, alignment strategies varied from zero-shot prompting and synthetic data generation to reinforcement and post-training fusion, yet they converged on preserving relevance and minimizing topic drift through principled model adaptation.

\subsection{Findings}

\subsubsection{RQ1}

Regarding \textbf{RQ1}, the reviewed methods achieved alignment to conversation goals using a range of strategies that included in-context imitation, token-level optimization, contrastive fine-tuning, and reinforcement learning. Prompt-based methods like \texttt{uril} operated without parameter updates, aligning responses through static instruction formats and stylistic exemplars. 
Specifically, \texttt{uril} achieved alignment by employing a static prompt structure and in-context learning that stylistically imitated aligned behavior, effectively guiding the model’s responses without requiring fine-tuning or reinforcement learning. 
Inference-time strategies such as \texttt{rain}, \texttt{args}, and Linear Alignment directly modified the decoding process to guide token selection based on self-evaluation or preference signals. 
For instance, \texttt{rain} integrated alignment into token selection via a self-evaluation and rewind mechanism during inference, enabling dynamic correction of outputs according to human-aligned goals.
Similarly, \texttt{args} applied alignment post hoc in decoding by adjusting token probabilities with a reward-guided scoring function trained on human preference data, while Linear Alignment used a closed-form update to token logits based on principle prompts to steer generation without external supervision or parameter changes. 
In contrast, post-training approaches like \texttt{dlma}, \texttt{rrhf}, and CodecLM fine-tuned models using preference-aware datasets, either synthetically generated or obtained from comparative scoring. 
\texttt{dlma} exemplified this by employing fine-tuning with self-generated contrastive prompts and internal scoring to optimize alignment, thus bridging inference-only and supervised approaches. Additionally, reinforcement-based methods, including PPO variants, Nash equilibrium learning, and principle-driven feedback, adjusted policies using reward signals that prioritized responses preferred under human or derived constraints. 
Across all approaches, alignment emerged through structured guidance applied at different points in the model lifecycle, with varying implications for efficiency and control.

\subsubsection{RQ2}

The reviewed studies addressed \textbf{RQ2} by grounding model outputs to ensure coherence, factuality, and thematic relevance through diverse mechanisms including prompt design, structured data generation, reward modeling, and architectural constraints. Prompt-based methods like \texttt{uril} and CodecLM encoded expected behavior formats and response constraints directly into the input, thereby providing a stable context that promoted well-structured and socially responsible outputs. For example, \texttt{uril} explicitly embedded expected response formats and behavioral guidelines in the system prompt and examples, which grounded generation effectively. Inference-time frameworks such as \texttt{rain} and \texttt{args} operationalized grounding by applying human-aligned evaluation criteria—such as harmlessness and factual accuracy—during decoding. \texttt{rain} used structured prompts to communicate these criteria and grounded generation through self-assessment and dynamic correction during inference, while \texttt{args} balanced language model likelihood with preference-based rewards to maintain coherence and incorporate human preferences. Similarly, Linear Alignment preserved semantic coherence by bounding update magnitudes to maintain the original output’s intent while steering generation according to principle-based prompts. Fine-tuning strategies including \texttt{dpo}, \texttt{rrhf}, \texttt{figa}, and \texttt{dlma} grounded behavior by leveraging contrastive or token-specific supervision that reflected contextual fidelity or factual correctness. \texttt{dlma} reinforced grounded behavior through instruction tuning and contrastive prompts emphasizing target attributes, ensuring the model prioritized helpful and safe completions. Reinforcement learning methods incorporated grounding by designing custom reward models, exemplified by retrieval-based entailment in \texttt{gpm}, correctness feedback in cRLHF, and principle alignment in SALMON. Additionally, KL-regularization featured prominently in multiple training-time methods to preserve proximity to initial instruction-tuned distributions, thereby maintaining thematic consistency with the input prompt and avoiding divergence from grounded baselines. Collectively, these approaches applied structured guidance at different stages of the model lifecycle to enforce grounded, coherent, and factually accurate outputs.

\subsubsection{RQ3}

The findings also provided evidence for how the reviewed alignment methods contributed to \textbf{RQ3}. Several techniques mitigated hallucinations and topic drift by constraining generation paths, selecting preferred completions, or penalizing unsafe continuations. Inference-time strategies like \texttt{rain}, \texttt{args}, and Linear Alignment implemented rejection or correction mechanisms that filtered low-reward tokens or adjusted outputs toward preference-consistent regions. Fine-tuning methods such as DLMA, RRHF, and CodecLM embedded quality constraints into training through self-evaluation, contrastive supervision, or synthetic filtering. Reinforcement-based approaches minimized hallucinations through reward shaping and constrained updates, with examples including KL-penalized policy optimization (PPO, RLOO), principle-driven reward modeling (SALMON), and entailment-based alignment (GPM). Across the reviewed literature, the combination of alignment mechanisms demonstrated convergent strategies for improving conversation fidelity while reducing unwanted output behaviors. Together, these findings clarified how current methods operationalized alignment along three axes—intent, grounding, and control—offering practical pathways for aligning large language models in both low- and high-resource settings.

\section{Discussion}\label{sec:discussion}
This discussion builds on the analysis presented in the results, where methods were categorized by the LLM lifecycle phase in which alignment was applied. The review examined how each method contributed to answering the three research questions (\textbf{RQ1}, \textbf{RQ2}, and \textbf{RQ3}) by focusing on alignment to conversational goals, support for grounding, and reduction of hallucinations. To complement this analysis, this section reinterprets the findings according to the type of alignment approach used. By stratifying the methods into reinforcement learning-based, direct optimization, self-alignment or AI feedback-based, and tuning-free inference strategies, it becomes possible to draw additional insights into how alignment objectives were operationalized and how different approaches balanced computational efficiency, feedback supervision, and training complexity.

Reinforcement learning methods, such as \texttt{ppo}, REINFORCE, \texttt{rloo}, and Nash equilibrium-based optimization, aligned model policies by optimizing reward-guided updates derived from preference models or task-specific signals. These methods aligned with \textbf{RQ1} by updating model policies to prefer high-reward responses, and supported \textbf{RQ2} through feedback grounded in retrieval evidence, preference comparisons, or task-specific goals. They addressed \textbf{RQ3} by constraining policy updates using KL-divergence or reward penalties to reduce hallucinations and topic drift. Alternative feedback mechanisms, including principle-based reward modeling in SALMON and contrastive signals from document clusters in RLCF, demonstrated that reinforcement learning could be extended beyond standard preference data without changing the core optimization strategy.

RL-free methods proposed alignment through direct optimization objectives, avoiding reward models and sampling loops. Techniques such as \texttt{dpo}, \texttt{rrhf}, and \texttt{figa} aligned models by optimizing on pairwise preferences or token-level corrections. These methods contributed to \textbf{RQ1} by updating parameters based on structured supervision and avoided reinforcement learning pipelines. They supported \textbf{RQ2} by reinforcing grounded behavior through contrastive supervision or revision-based data. For \textbf{RQ3}, they suppressed hallucinated outputs by down-weighting low-preference completions. Some approaches, such as ARM and RS-DPO, combined reward-based filtering with offline optimization, replicating aspects of RLHF without requiring online sampling or gradient-based reward modeling.

Several methods used self-generated or AI-generated feedback to approximate human preference supervision. These approaches aligned models by employing internal evaluation signals, contrastive prompt structures, or self-scored outputs. DLMA and Self-Alignment for Factuality applied internal scoring mechanisms to rank responses, while SALMON trained instructable reward models from principle-based inputs. These strategies addressed \textbf{RQ1} through preference modeling without external annotation, and supported \textbf{RQ2} by grounding outputs using internally consistent criteria or synthetic structures. They addressed \textbf{RQ3} by selecting or reinforcing high-confidence responses, rejecting outputs with low internal evaluations. These methods reduced dependency on labeled data and introduced scalable alternatives to traditional supervision.

Inference-time methods aligned language models without modifying parameters by guiding generation through decoding algorithms or prompt engineering. \texttt{uril} aligned behavior using fixed prompts and stylistic examples, relying on in-context learning. \texttt{rain} implemented a rewind mechanism for dynamic token correction, while \texttt{args} adjusted token probabilities using a reward-guided scoring function. Linear Alignment introduced a closed-form update to token logits using self-contrastive decoding with principle prompts. These methods supported \textbf{RQ2} by embedding constraints into prompts or decoding, and addressed \textbf{RQ3} by rejecting or revising low-alignment continuations. They differed in implementation but converged on the strategy of applying alignment during generation, not through training or parameter changes.

The comparative findings indicated that inference-time methods consistently required fewer computational resources than training-based techniques while maintaining effective alignment. Among these, URIAL demonstrated that prompt engineering alone could induce stylistic and safety alignment in large frozen models. RAIN enabled alignment via runtime self-evaluation and correction, while Linear Alignment produced decoding adjustments based on principle prompts through a one-step update.
ARGS integrated reward modeling into token-level generation decisions, aligning outputs without backpropagation or additional training cycles. These methods differed in whether alignment emerged from stylistic imitation, scoring-based token selection, or contrastive updates, but all shared the characteristic of aligning behavior during generation rather than through parameter updates. 
Their efficiency and adaptability suggest that inference-time strategies represent a viable direction for low-resource or real-time alignment use cases.

\section{Conclusion} \label{sec:conclusion}
In this review, we aimed to investigate the methods and approaches that increased the LLM's ability to provide alignment during conversation, enabling grounded responses and reducing topic deviation and hallucination in this context. According to the analysis, several factors contributed to LLM alignment. This analysis was stratified into three main phases of the LLM life cycle: inference-time, post-training, and reinforcement learning-based models.

Each of these phases addressed specific alignment strategies. In particular, the inference-time methods examined in this review consistently targeted the alignment of LLM outputs with user intent and contextual goals, while circumventing the need for parameter updates. These approaches primarily operated by manipulating decoding strategies during inference, leveraging external or internal signals of preference. Some relied on reward models to guide generation, either by dynamically evaluating candidate outputs or by influencing token-level probabilities—without engaging in full reinforcement learning loops. Others incorporated self-evaluation and feedback loops into the generation process, allowing the model to revise or optimize outputs mid-decoding. To reduce computational demands, these methods avoided iterative fine-tuning and instead employed lightweight techniques such as contrastive comparisons, scoring-based selection, or structured prompt strategies. In cases where preference optimization was approximated, the solution was often derived in closed form at decoding time, eliminating the need for gradient-based learning. Across these techniques, prompt engineering played a supportive rather than central role, often used to elicit initial structure or behavior, while alignment was achieved through decoding-level control or sampling-based filtering. These strategies reflected a broader shift toward efficient alignment mechanisms that decoupled response quality from training cost.

Implications of these insights extend to developers and researchers involved in the development of inference-time alignment methods, as this study consolidated existing strategies that operated without fine-tuning and demonstrated their relevance for achieving alignment with reduced computational requirements. This study also encountered some limitations, such as the number of papers analyzed. As a future step to enhance the conclusions, it would be appropriate to apply the snowballing technique to retrieve additional articles based on those already selected in the filtering process. Another possible refinement would involve analyzing the responses to the research questions by stratifying the techniques according to the type of alignment employed (e.g., through AI feedback, human feedback, or in-context learning). This would offer more detailed insights into the conditions under which each alignment technique may be applied effectively.

\printbibliography 

\end{document}